\pdfoutput=1
\def\year{2020}\relax
\documentclass[letterpaper]{article} 
\usepackage{aaai20}  
\usepackage{times}  
\usepackage{helvet} 
\usepackage{courier}  
\usepackage[hyphens]{url}  
\usepackage{graphicx} 
\usepackage{multirow}
\usepackage{enumerate} 
\usepackage{amssymb}
\usepackage{amsmath}
\usepackage{algorithm}
\usepackage{algorithmic}
\usepackage{mathtools}
\usepackage{multicol}
\def\UrlFont{\rm}  
\usepackage{graphicx}  
\title{Learning Continually from Low-shot Data Stream }
\author{Canyu Le\textsuperscript{\rm 1}\textsuperscript{\rm 2} \thanks{Part of this work were done when Canyu Le interned at Artifical Intelligence Center, DAMO Academy, Alibaba Group.}, 
	Xihan Wei\textsuperscript{\rm 2}, 
	Biao Wang\textsuperscript{\rm 2}, 
	Lei Zhang\textsuperscript{\rm 2},
	Zhonggui Chen\textsuperscript{\rm 1}\\
	\textsuperscript{\rm 1} Department of Computer Science, Xiamen University\\
	\textsuperscript{\rm 2} DAMO Academy, Alibaba Group.\\
	\texttt{lecanyu@gmail.com}  \quad \texttt{\string{xihan.wxh, wb.wangbiao, lei.zhang.lz\string}@alibaba-inc.com}
}

\begin{document}
	
	\maketitle
	
	\begin{abstract}
		While deep learning has achieved remarkable results on various applications, it is usually data hungry and struggles to learn over non-stationary data stream. 
		To solve these two limits, the deep learning model should not only be able to learn from a few of data, but also incrementally learn new concepts from data stream over time without forgetting the previous knowledge.  
		Limited literature simultaneously address both problems.
		In this work, we propose a novel approach, MetaCL, which enables neural networks to effectively learn meta knowledge from low-shot data stream without catastrophic forgetting. 
		MetaCL trains a model to exploit the intrinsic feature of data (i.e. meta knowledge) and dynamically penalize the important model parameters change to preserve learned knowledge.
		In this way, the deep learning model can efficiently obtain new knowledge from small volume of data and still keep high performance on previous tasks.  
		MetaCL is conceptually simple, easy to implement and model-agnostic.
		We implement our method on three recent regularization-based methods. 
		Extensive experiments show that our approach leads to state-of-the-art performance on image classification benchmarks.
	\end{abstract}
	
	\section{Introduction}
\begin{figure}[h]
		\centering
		\includegraphics[width=0.4\textheight]{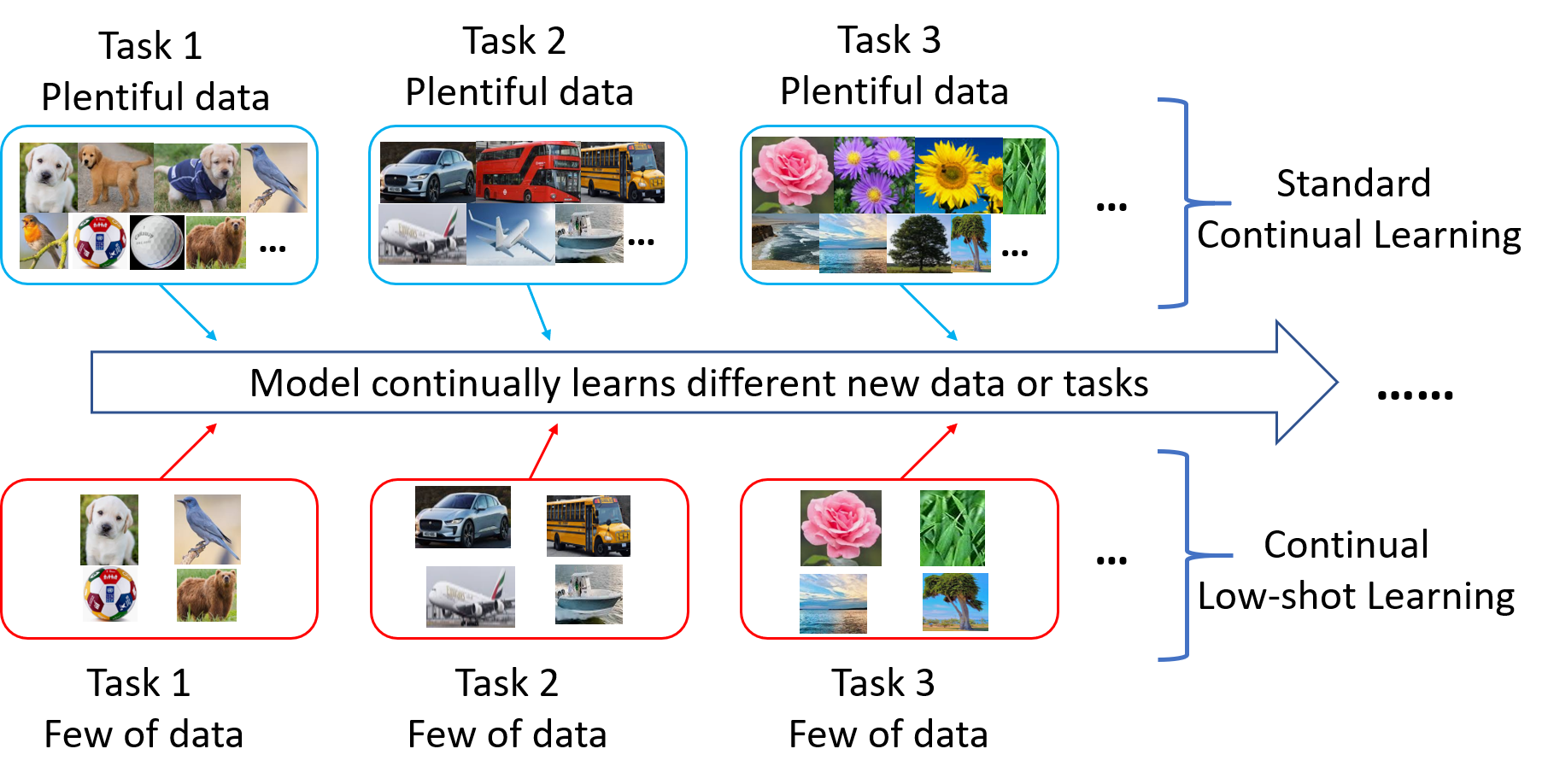}
		\caption{The comparison between standard continual learning and continual low-shot learning for image classification. 
			The top row is standard CL in which each task has plentiful training data.
			The bottom row is continual low-shot learning where only a handful of training data for each task.  \label{Fig:continual_low_shot}}
	\end{figure}

	Human-level intelligence has two remarkable hallmarks: quick learning and slow forgetting. Human can efficiently learn to recognize new concepts from a few of examples without forgetting the prior knowledge. Ideally, the artificial agent should be able to demonstrate the same capabilities, learning continually from small volume of data and preserving what it has learned. 
	We call this human-like learning scenario as \emph{continual low-shot learning}, which can be seen as a generalization of the standard \emph{continual learning} (CL) \cite{li2017learning,rebuffi2017icarl}. 
	The comparison between the standard CL and continual low-shot learning is illustrated in Figure \ref{Fig:continual_low_shot}.
	The characteristics of continual low-shot learning problem can be formulated as follows: 
	\begin{enumerate}[(i)]
		\item Non-stationary data. A model will be trained in the whole data stream where new task data become available at different phases. 
		Compared with the previous tasks, the new task data could have different data distribution and categories.  
		\item Efficiency. During training and testing, the system resource consumption and computational complexity should be bounded. For example, when model learns new tasks, it cannot see old task data for quick learning and storage saving.
		\item Small size of data. The volume of training samples could be small (e.g. a few or dozens of training data).  
	\end{enumerate}
	The first two criteria are the important properties of the standard continual learning. The third criterion generalizes CL to address low-shot learning. 
	This generalization is important in many practical scenarios.
	For example, in realistic vision applications (e.g. classification, detection), the labeled training data is usually rare and can only be available incrementally due to high cost data labeling.
	It could be beneficial that a model can effectively learn from a small size of data and continually evolve itself as new data are available.
	Despite its importance, limited literature discussed this practical and more human-like learning problem.
	
	In continual low-shot learning, a model should demonstrate good performance in the entire data stream where the volume of each task is small. 
	Hence, learning efficiently from limited training data and simultaneously preserving learned knowledge are crucial. 
	Efficient learning means that a model can quickly learn the intrinsic knowledge from the limited data and obtain generalization.
	Knowledge preservation entails that new data learning should not cause negative interference in learned knowledge.
	The interference, however, is inevitable since the architecture of deep learning model is highly coupled. 
	Paucity of old task data supervision, the new data learning usually cause severe negative interference and the performance on previous tasks quickly deteriorates, which is so-called \emph{catastrophic forgetting} \cite{mccloskey1989catastrophic}.
	These two properties, efficient learning and knowledge preservation, usually conflict with each other and it is challenging to find the optimal trade-off.
	
	In this work, we propose a novel algorithm to address this challenge from two aspects. 
	(1) In contrast to prior methods which focus on how to reduce forgetting \cite{zenke2017continual,kirkpatrick2017overcoming,aljundi2018memory}, we try to strengthen model adaptation via a multi-steps optimization procedure. 
	This procedure can efficiently learn meta knowledge from a small size of data, and the strong adaptation can also give more potential space for learning-forgetting compromise.
	(2) Instead of applying a fixed hyperparameter to balance learning objective and regularization terms, we develop a dynamic balance strategy by altering optimization gradients.   
	This dynamic strategy provides a comparable or better trade-off between learning and forgetting, and thus further improves the overall performance. 
	
	For knowledge preservation, we adopt the parameter regularization based approaches, which measures the importance of model parameters and penalize its change in new task training. 
	Compared with other approaches like model expansion \cite{aljundi2017expert,rusu2016progressive} and gradient regularization \cite{lopez2017gradient,chaudhry2018efficient}, the parameter regularization is more computational efficient and does not access previous task data. 
	
	We implement our model-agnostic algorithm MetaCL based on three state-of-the-art parameter regularization methods: EWC \cite{kirkpatrick2017overcoming}, PI \cite{zenke2017continual}, and MAS \cite{aljundi2018memory}. 
	And extensive experiments show that our approach can further improve those baselines.

	In summary, our main contributions of this work include
	\begin{itemize}
		\item We design a model-agnostic algorithm, MetaCL, which strengthen model adaptation ability in continual low-shot learning without using any data in previous tasks.
		\item We develop a dynamic balance strategy to adaptively penalize parameter changes to stabilize optimization gradients and achieve better trade-off between current task learning and previous task forgetting. 
		\item We compare our approach with existing algorithms under various experimental settings and analyze them in terms of accuracy, forgetting, and adaptation. 
	\end{itemize}

	\section{Related Work}
	Our approach builds on the insights of model adaptation and knowledge preservation. These two characteristics have been mainly addressed in meta learning and continual learning fields.  
	We briefly discuss both.
	
	\textbf{Meta learning}. 
	The main goal in meta learning is to endow a model with strong adaptation ability, so as to a model trained on a domain (i.e. so-called meta training dataset) can be quickly transferred to other new domains (i.e. meta testing dataset) where only few of labeled data (i.e. support set) are available.
	Generally, the existing methods can be categorized into three categories: metric-based, model-based and optimization-based.
	Metric-based approaches \cite{vinyals2016matching,snell2017prototypical,sung2018learning} try to learn a similarity metric so that the model can obtain more general and intrinsic knowledge.
	Model-based approaches \cite{santoro2016meta,munkhdalai2017meta} achieve adaptation via altering model components. 
	Optimization-based methods \cite{finn2017model} apply new optimization algorithms to find a good initialization. 
	However, all above approaches only consider how to learn from few-shot data, regardless of the model knowledge preservation.
	More recently, \cite{gidaris2018dynamic} implemented a meta-learning model through a similarity-based classifier and weight generator. 
	It protects the performance on meta training dataset after fine tuning on support set.
	\textbf{Nevertheless, our continual low-shot learning differs from meta learning in two significant aspects.
		First, there is no extra dataset (i.e. meta training dataset) for prior knowledge obtaining in continual low-shot learning.
		Second, instead of only two different datasets/tasks, the model faces theoretically unlimited tasks in continual low-shot learning}.
	So the existing meta learning methods cannot be directly applied to solve our problem. 
	
	\textbf{Continual learning}, on the other hand, mainly focuses on how to remedy the catastrophic forgetting when model learns new tasks.
	Most existing literature addressed this problem from two aspects: model decoupling and model regularization. 
	\cite{aljundi2017expert,aljundi2018selfless} decouple model to decrease the interference when learning new data.
	Model regularization methods \cite{li2017learning,kirkpatrick2017overcoming} add an extra regularization term to preserve learned knowledge. 
	In spite of their effectiveness in knowledge preservation, these methods neglect the low-shot scenarios and adaptation ability.
	Later, \cite{lopez2017gradient,chaudhry2018riemannian} observed the compromise between learning and forgetting.
	But they didn't develop a strategy to explicitly enhance learning and adaptation ability.  
	
	In contrast to prior methods, we address continual low-shot learning and propose a model-agnostic algorithm that strengthens adaptation and provides a better trade-off between learning and forgetting.
	Our method neither modifies the network architecture nor relies on external experience memory. This makes our method memory efficient and easy to be extended to other existing models and applications.
	
	\section{Approach}
	We aim to train a model to obtain strong adaptation and preserve its performance on previous tasks.   
	In the following, we will define the problem setup and present our approach in classification context, but the idea can be extended to other learning problems.
	
	\subsection{Continual Low-shot Learning Problem Setup}
	The goal of continual low-shot learning is to train a model that can not only quickly adapt to a new task using a small size of data but also demonstrate high performance on previous tasks.
	In particular, the model $f_\theta$, which is parameterized by $\theta \in \mathbb{R}^p$ will be trained on a stream of data $(x_i, y_i, t_j)$, where the $t_j \in \mathcal{T} (j=1,2,...,n)$ is the task descriptor and $(x_i, y_i) \in \mathcal{X}_j$ is a data point in task $j$.
	In continual low-shot learning, the volume of training data for each task is small. 
	Besides, the model $f_\theta$ can only see the training dataset $\mathcal{X}_j$ when learning task $j$.
	Formally, the objective function can be written as:
	\begin{equation}
	\label{Eqn:obj}
	\min_\theta L(f_\theta, \mathcal{X}, \mathcal{T}) = \sum_{t_j \in \mathcal{T}} \sum_{(x_i, y_i) \in \mathcal{X}_j} \ell(f_\theta(x_i, t_j), y_i)
	\end{equation}
	where $\ell(\cdot, \cdot)$ is the loss function which could be cross-entropy in image classification.
	For simplicity, we will use $\ell(\theta)$ to denote $\ell(f_\theta(x_i, t_j), y_i)$ in the following formulations.
	
	If all task data are available in one training phase, we can trivially train all data to minimize above objective Eq. \ref{Eqn:obj} (a.k.a. \emph{joint training}).
	In continual low-shot learning, however, only current task data can be accessed during a training stage.
	Under such incomplete supervision, the model is prone to encounter catastrophic forgetting.
	
	\subsection{Reducing Forgetting}
	To alleviate the forgetting problem, we adopt parameter regularization-based methods which measures the parameter importance in prior tasks and penalizes its change in new task training. 
	As indicated in \cite{chaudhry2018riemannian}, this kind of method is more memory efficient and scalable than activation (output) regularization \cite{rebuffi2017icarl,li2017learning} and network expansion methods \cite{yoon2018lifelong,rusu2016progressive,aljundi2017expert}.
	
	Generally, the parameter regularization for learning task $t_j$ can be formulated as below:
	\begin{equation}
	\label{Eqn:final_obj}
	L_{t_j} = \sum_{(x_i, y_i) \in \mathcal{X}_j} [\ell(f_\theta(x_i, t_j), y_i) + \beta \sum_{k=1}^p \Omega_k (\theta_k - \bar{\theta}_k)^2 ]
	\end{equation}
	where $\Omega_k$ is the importance measure for $k$-th parameter $\theta_k$ (total $p$ parameters in model). $\bar{\theta}_k$ is the pretrained parameter from previous tasks $t_1, t_2, ..., t_{j-1}$. $\beta$ is a hyperparameter which balance current task $j$ learning and previous tasks forgetting. Obviously, the bigger $\beta$ is, the stronger knowledge preservation and less knowledge update can be achieved. 
	
	
	There are two key problems in parameter regularization: (1) how to calculate the importance measure $\Omega_k$ and (2) how to set a proper hyperparameter $\beta$ to get a good trade-off. 
	A lot of literature \cite{lee2017overcoming,zenke2017continual,aljundi2018memory,chaudhry2018riemannian} have addressed the first problem, but few discuss the second one.
	In this work, we develop a dynamic balance strategy that address the latter problem.

	\subsection{Dynamic Balance Strategy}
	There are two terms for every data point optimization in Eq. \ref{Eqn:final_obj}.
	The first term $\ell(\theta) \coloneqq \ell(f_\theta(x_i, t_j), y_i)$ drives the model toward current task learning.
	The second regularization term $ \ell^{reg}(\theta) \coloneqq \sum_{k=1}^p \Omega_k (\theta_k - \bar{\theta}_k)^2$ preserves the previous task knowledge.
	A fixed hyperparameter $\beta$ is applied to balance current task learning and old knowledge preservation.
	This simple balance strategy is widely adopted in many existing model regularization methods like \cite{zenke2017continual,kirkpatrick2017overcoming,aljundi2018memory}.
	However, one has to spend a lot of time to manually search a proper hyperparameter. 
	Besides, if the gradients of those two terms are unstable, the fixed hyperparameter may not be able to provide a good compromise between $\ell(\theta)$ and $\ell^{reg}(\theta)$ in the entire data stream (a concrete example is given in Experiment Section).
	
	To mitigate these problems, we propose a dynamic balance strategy which adaptively adjusts the gradient direction to compromise current task learning and knowledge preservation.
	The key intuition behind this strategy is that a good balance can be reached if we can find an optimization direction $g_x$ which satisfies the following two conditions:
	(1) $g_x$ is as close as possible to the gradient of current task learning  $g_1 = \frac{\partial \ell(\theta)}{\partial \theta}$;
	(2) optimizing along with $g_x$ should not increase the second regularization term $\ell^{reg}$ for knowledge preservation. 
	
	Suppose the objective function is locally linear (it happens around small optimization steps), we can formulate above intuition in a constrained optimization problem:
	\begin{align}
	\label{Eqn:Dynamic_balancing_obj}
	\min_{g_x} \frac{1}{2} \|g_x - g_1 \|^2 \notag \\
	s.t. \quad \langle g_x, g_2 \rangle \ge 0
	\end{align}
	where $g_2 = \frac{\partial \ell^{reg}(\theta)}{\partial \theta}$, the operator $\langle \cdot, \cdot \rangle$ is dot product. 
	The optimization object in Eq. \ref{Eqn:Dynamic_balancing_obj} indicates that the $g_x$ should be as close as possible to $g_1$ in the squared $\ell_2$ norm. 
	The constraint term represents that the gradient angle between $g_x$ and $g_2$ should be smaller than $90^{\circ}$ so that the optimization toward $g_x$ doesn't increase the second regularization term $\ell^{reg}$.
	Since $g_x$ has $p$ variables (the number of parameters in the neural network), it is intractable to solve Eq. \ref{Eqn:Dynamic_balancing_obj} directly.
	We apply the principle of quadratic program and its dual problem \cite{dorn1960duality}, and the Eq. \ref{Eqn:Dynamic_balancing_obj} can be converted to its dual space (please check Appendix A for detailed derivation):
	\begin{align}
	\label{Eqn:Dynamic_balancing_dual_obj}
	\min_{\lambda} \frac{1}{2} & g_2^T g_2 \lambda^2 + g_1^T g_2 \lambda  \notag \\
	s.t. \quad & \lambda \ge 0,  \notag \\
	& g_x = \lambda g_2 + g_1 
	\end{align}
	where $\lambda$ is a Lagrange multiplier.
	
	Eq. \ref{Eqn:Dynamic_balancing_dual_obj} is a simple one-variable quadratic optimization.
	The optimal $\lambda$ is 
	\begin{align}
	\label{Eqn:Dynamic_balancing_lambda_solution}
	\lambda =
	\begin{cases}
	0 & \mbox{if $ g_1^T g_2 \ge 0 $}\\
	- \frac{g_1^T g_2}{g_2^T g_2} & \mbox{if $  g_1^T g_2 < 0 $}
	\end{cases}
	\end{align}
	Then, we can calculate the optimal $g_x =  g_1 + \lambda g_2$. 
	
\begin{figure}[h]
		\centering
		\begin{tabular}{c}
			\includegraphics[width=0.35\textheight]{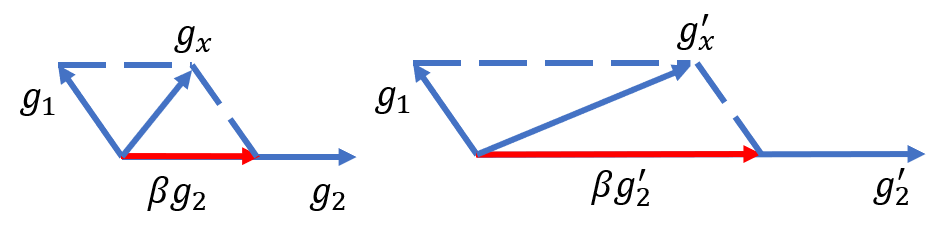} \\
			(a) Fixed balance strategy. $\beta$ is a fixed fraction (e.g. 0.6). \\
			\includegraphics[width=0.35\textheight]{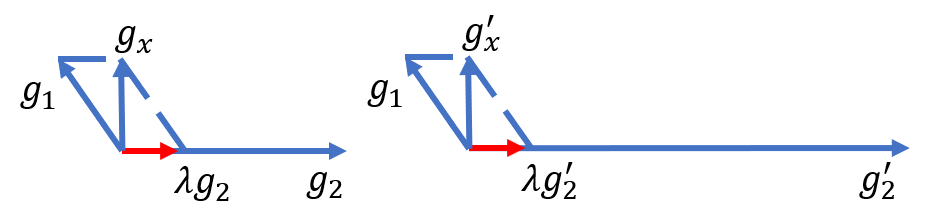} \\
			(b) Dynamic balance strategy. $\lambda$ is dynamically determined.
		\end{tabular}
		\caption{The difference between two strategies. The dynamic balance strategy can provide more reliable optimization direction $g_x$, even though $g_2$ grows in optimization procedure. \label{Fig:dynamic}}
	\end{figure}

	As a comparison, the gradient in fixed balance strategy is $g = g_1 + \beta g_2$, 
	whereas the dynamic balance strategy uses the gradient $g_x = g_1 + \lambda g_2$ with the adaptive weight $\lambda= -\frac{g_1^T g_2}{g_2^T g_2}$.
	Fig. \ref{Fig:dynamic} shows the difference between two strategies.
	Since $g_1, g_2$ are related with current parameters and data point, $\lambda$ can vary and adaptively balance $\ell(\theta)$ and $\ell^{reg}(\theta)$ during the whole of training procedure.
	In practice, we found that adding a small constant $\gamma > 0$ to the adaptive weight $\lambda$ will further fortify the knowledge preservation.
	
	\subsection{Strengthening Adaptation}
	If there are sufficient training data in task $j$, we may directly train a model based on Eq. \ref{Eqn:final_obj} and achieve desirable results.
	But this assumption doesn't hold in continual low-shot learning problem where the size of training data for a task is small.
	To address this low-shot learning problem, the model needs to adequately exploit the intrinsic features from limited data.
	One way to do so is to maximize the inner product between gradients of different data points within a task:
	\begin{equation}
	\label{Eqn:inner_g}
	\max \frac{\partial \ell(f_\theta(x_u, t_j), y_u)}{\partial \theta} \cdot \frac{\partial \ell(f_\theta(x_v, t_j), y_v)}{\partial \theta}
	\end{equation}
	Eq. \ref{Eqn:inner_g} can lead the learning procedure to find common features among different data rather than just fitting a single data point.
	
	Combining Eq. \ref{Eqn:inner_g} and Eq. \ref{Eqn:final_obj}, we are interested to optimize the below new objective: 
	\begin{equation}
	\label{Eqn:new_obj}
	L_{t_j} = \sum_{u, v \in \mathcal{X}_j} [\ell_u(\theta) + \ell_v(\theta) - \alpha \frac{\partial \ell_u(\theta)}{\partial \theta} \cdot \frac{\partial \ell_v(\theta)}{\partial \theta} + \beta \ell^{reg}(\theta)]
	\end{equation}
	where $\ell_u(\theta), \ell_v(\theta)$ denote the losses at data points $(x_u, y_u), (x_v, y_v)$ respectively.
	Optimizing Eq. \ref{Eqn:new_obj} needs the second derivative w.r.t. $\theta$, which is expensive to calculate.
	Inspired from the recent meta-learning algorithm, Reptile \cite{nichol2018first}, we can design a multi-step optimization algorithm that bypasses the second derivative calculation and seamlessly integrates with parameter importance measurement.
	The complete MetaCL is outlined in Algorithm \ref{Alg:MetaCL_beta}.
	
	\begin{algorithm}  
		\caption{MetaCL-$\beta$ (fixed balance version)}  
		\label{Alg:MetaCL_beta}  
		\begin{algorithmic}  
			\REQUIRE{The training data $\mathcal{X}_j$ in task $t_j$, the model $f$ with pretrained parameter $\bar{\theta}$. Step size hyperparameters $\alpha, \eta$. Balance hyperparameter $\beta$.}
			\ENSURE{The new model parameter $\theta^*$}
			\STATE $f_{\theta} \gets$ load the pretrained parameter $\bar{\theta}$.
			\FOR{ epoch$=1, 2, ...$}
			\FOR{ mini-batch $B$ in $\mathcal{X}_j$ } 
			\item Randomly split mini-batch $B$ to mini-bundles $b_1, b_2, ..., b_m$.
			\item // Inner loop optimization.
			\FOR{ $i=1,2,...,m$}
			\item $\theta^{i} = \theta^{i-1} - \alpha \ell_{b_i}^{'}(\theta^{i-1}) $. (Note that $\theta^{0} \equiv \theta$)
			\ENDFOR
			\item // The gradient for current task learning
			\item $g_1 = (\theta - \theta^{m})/(\alpha * m)$
			\item // The gradient for forgetting reducing
			\item $g_2 = \ell^{reg '}(\theta)$
			\item Calculate $g = g_1 + \beta g_2$
			\item Update $\theta \gets \theta - \eta * g$
			\ENDFOR
			\ENDFOR
			\STATE $\theta^* = \theta$
		\end{algorithmic}  
	\end{algorithm}
	
	\textbf{Algorithm analysis}.
	Algorithm \ref{Alg:MetaCL_beta} implicitly satisfies the objective Eq. \ref{Eqn:new_obj}.
	Let's check the current task learning gradient $g_1$ to explain how it works.
	
	If we sum up all mini-bundles optimization in the inner loop of Algorithm \ref{Alg:MetaCL_beta}, we have
	\begin{equation}
	\theta^0 - \theta^m = \theta - \theta^m = \alpha \sum_{i=1}^{m} \ell_{b_i}^{'}(\theta^{i-1})
	\end{equation} 
	Therefore, the gradient $g_1$ can be rewritten as:
	\begin{equation}
	\label{Eqn:g_1}
	g_1 = \frac{\theta-\theta^m}{\alpha m} = \frac{1}{m} \sum_{i=1}^{m} \ell_{b_i}^{'}(\theta^{i-1})
	\end{equation}
	By applying Taylor series expansion on $\ell_{b_i}^{'}(\theta^{i-1})$, we have  
	
	\begin{align}
	\label{Eqn:taylor_exp}
	\ell_{b_i}^{'}(\theta^{i-1}) &= \ell_{b_i}^{'}(\theta^{0}) + \ell_{b_i}^{''}(\theta^{0})(\theta^{i-1} - \theta^0) + O(\Vert \theta^{i-1} - \theta^0\Vert^2) \notag \\
	&\approx \ell_{b_i}^{'}(\theta) + \ell_{b_i}^{''}(\theta)(\theta^{i-1} - \theta^0) \notag \\
	&= \ell_{b_i}^{'}(\theta) - \alpha \ell_{b_i}^{''}(\theta) \sum_{k=1}^{i-1} \ell_{b_k}^{'}(\theta^{k-1}) \notag \\
	\end{align}
	Apply Taylor series expansion on $\ell_{b_k}^{'}(\theta^{k-1})$ again:
	\begin{equation}
	\label{Eqn:taylor_exp2}
	\ell_{b_k}^{'}(\theta^{k-1}) = \ell_{b_k}^{'}(\theta^0) + O(\Vert \theta^{i-1} - \theta^0\Vert) \approx \ell_{b_k}^{'}(\theta)
	\end{equation}
	These approximation can hold if the $m, \alpha$ are small (i.e. small update in inner loop optimization).
	
	Substituting Eq. \ref{Eqn:taylor_exp2} into Eq. \ref{Eqn:taylor_exp}, we have:
	\begin{equation}
	\label{Eqn:b_i}
	\ell_{b_i}^{'}(\theta^{i-1}) \approx \ell_{b_i}^{'}(\theta) - \alpha \ell_{b_i}^{''}(\theta) \sum_{k=1}^{i-1} \ell_{b_k}^{'}(\theta)
	\end{equation}
	Since the mini-batches and mini-bundles are randomly sampled, the data point subscript exchange should be satisfied: $\ell_{b_i}^{''}(\theta)\ell_{b_k}^{'}(\theta) = \ell_{b_k}^{''}(\theta)\ell_{b_i}^{'}(\theta)$.
	Therefore, the Eq. \ref{Eqn:b_i} can be converted to
	\begin{align}
	\label{Eqn:b_i_final}
	\ell_{b_i}^{'}(\theta^{i-1}) &\approx \ell_{b_i}^{'}(\theta) - \alpha \ell_{b_i}^{''}(\theta) \sum_{k=1}^{i-1} \ell_{b_k}^{'}(\theta) \notag \\
	&= \ell_{b_i}^{'}(\theta) - \frac{1}{2}\alpha \sum_{k=1}^{i-1} (\ell_{b_i}^{''}(\theta)\ell_{b_k}^{'}(\theta) + \ell_{b_k}^{''}(\theta)\ell_{b_i}^{'}(\theta) ) \notag \\
	&= \ell_{b_i}^{'}(\theta) - \frac{1}{2}\alpha \sum_{k=1}^{i-1} \frac{\partial \ell_{b_i}^{'}(\theta)\ell_{b_k}^{'}(\theta) }{\partial \theta}
	\end{align}
	Substituting Eq. \ref{Eqn:b_i_final} into Eq. \ref{Eqn:g_1}, we can see
	\begin{equation}
	g_1 = \frac{1}{m} \sum_{i=1}^{m} [ \ell_{b_i}^{'}(\theta) - \frac{1}{2}\alpha \sum_{k=1}^{i-1} \frac{\partial \ell_{b_i}^{'}(\theta)\ell_{b_k}^{'}(\theta) }{\partial \theta} ]
	\end{equation}
	$\ell_{b_i}^{'}(\theta)$ is the gradient to minimize the loss at mini-bundle $b_i$.
	The second term $\sum_{k=1}^{i-1} \frac{\partial \ell_{b_i}^{'}(\theta)\ell_{b_t}^{'}(\theta) }{\partial \theta}$ is the inner product between gradients of different mini-bundles.
	It indicates that the model should be optimized to not only fit current mini-bundle but also learn the common features among different mini-bundles.
	The common feature learning, which can be seen as meta knowledge, strengthens adaption and generalization.
	When $m=2$, the $g_1$ can be seen as the gradient for current task learning in objective Eq. \ref{Eqn:new_obj}.
	
	As explained in the previous subsection, the fixed balance strategy may cause several problems and dynamic balance is more desirable when optimization gradients are unstable.  
	We integrate this dynamic balance strategy to our MetaCL algorithm, called MetaCL-$\lambda$, which is concluded in Algorithm \ref{Alg:MetaCL_lambda}.
	\begin{algorithm}  
		\caption{MetaCL-$\lambda$ (dynamic balance version)}  
		\label{Alg:MetaCL_lambda}  
		\begin{algorithmic}  
			\REQUIRE{The training data $\mathcal{X}_j$ in task $t_j$, the model $f$ with pretrained parameter $\bar{\theta}$. Step size hyperparameters $\alpha, \eta$.}
			\ENSURE{The new model parameter $\theta^*$}
			\STATE $f_{\theta} \gets$ load the pretrained parameter $\bar{\theta}$.
			\FOR{ epoch$=1, 2, ...$}
			\FOR{ mini-batch $B$ in $\mathcal{X}_j$ } 
			\item Randomly split mini-batch $B$ to mini-bundles $b_1, b_2, ..., b_m$.
			\item // Inner loop optimization.
			\FOR{ $i=1,2,...,m$}
			\item $\theta^{i} = \theta^{i-1} - \alpha \ell_{b_i}^{'}(\theta^{i-1}) $. (Note that $\theta^{0} \equiv \theta$)
			\ENDFOR
			\item // The gradient for current task learning
			\item $g_1 = (\theta - \theta^{m})/(\alpha * m)$
			\item // The gradient for forgetting reducing
			\item $g_2 = \ell^{reg '}(\theta)$
			\item Calculate $\lambda$ using Eq. \ref{Eqn:Dynamic_balancing_lambda_solution}.
			\item Calculate the optimization gradient $g_x =  g_1 + \lambda g_2$.
			\item Update $\theta \gets \theta - \eta * g_x$
			\ENDFOR
			\ENDFOR
			\STATE $\theta^* = \theta$
		\end{algorithmic}  
	\end{algorithm}

	\section{Experiments}
	We conduct experiments to evaluate baselines and our proposed MetaCL in various public benchmarks and settings.
	
	\subsection{Datasets}
	We use three datasets: \emph{Permuted MNIST} \cite{kirkpatrick2017overcoming}, \emph{CIFAR100} \cite{krizhevsky2009learning} and \emph{CUB} \cite{WahCUB_200_2011}.
	Permuted MNIST is a variant of the standard handwritten digits dataset, MNIST \cite{lecun1998mnist}, where the data in each task are arranged by a fixed permutation of pixels, and thus the data distribution between different tasks is unrelated.
	The CIFAR100 dataset contains 60k 32$\times$32 images with 100 different classes. 
	The CUB dataset has roughly 12k high resolution images with 200 fine-grained bird classes. 
	These datasets have been widely used in a variety of continual learning methods evaluation \cite{zenke2017continual,aljundi2018memory}.
	
	The size of original training datasets is large. 
	To simulate the low-shot setting, we sample the first $K$ images from each class to create a small volume of training data and use original testing data to evaluate.
	Note that when $K=1, 5$, the setting is similar with the 1-shot and 5-shot meta-learning \cite{finn2017model}.
	In contrast to meta learning, however, our continual low-shot learning problem does not have meta-training dataset to learn prior knowledge before learning consecutive task streams.
	We observe that there is no algorithm that can effectively learn from scratch without overfitting when $K=1, 5$.
	In this work, we typically sample $K=10, 20$, and put the extreme low-shot $K=1, 5$ for future study.
	
	\begin{figure*}[h]
		\centering
		\setlength\tabcolsep{0.5pt}
		\begin{tabular}{cccc}
			\includegraphics[width=0.19\textheight]{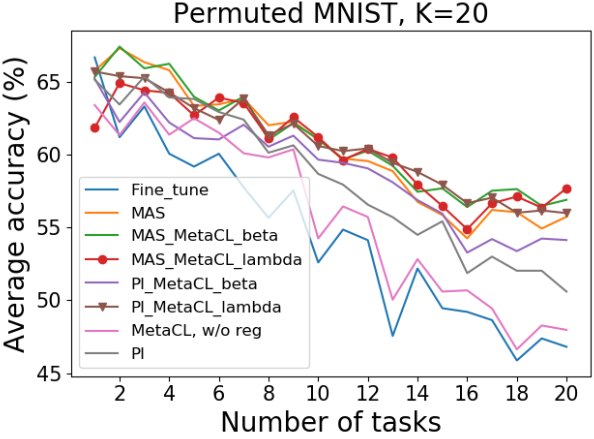} &
			\includegraphics[width=0.19\textheight]{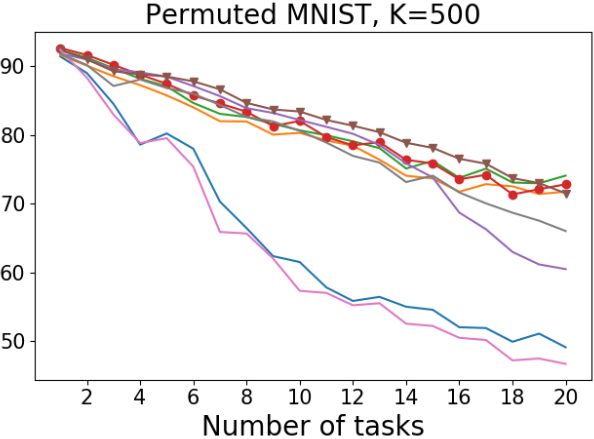} &
			\includegraphics[width=0.19\textheight]{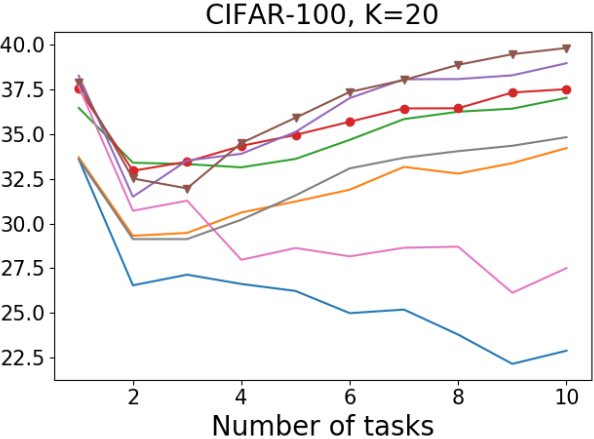} &
			\includegraphics[width=0.19\textheight]{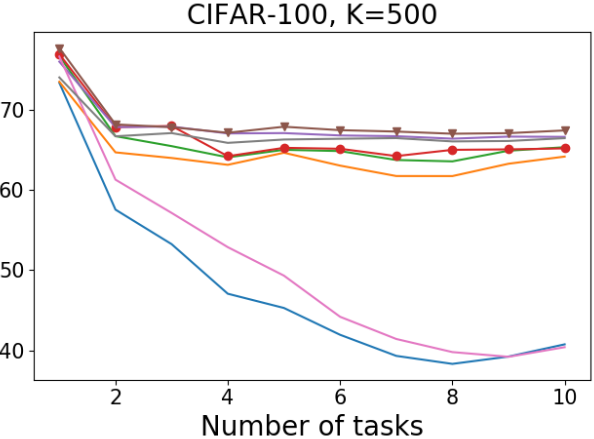}
		\end{tabular}
		\caption{The average accuracy changes as more tasks are learned with different K. The parameter regularization based methods relieve knowledge forgetting and MetaCL algorithm can further improve the model performance,  especially on low-shot setting.  \label{Fig:average_acc}}
	\end{figure*}
	
	\subsection{Metrics}
	We use the following metrics to quantitatively evaluate:  
	
	\textbf{Average Accuracy (ACC)}: if we define $a_{i,j}$ as the testing accuracy on task $j$ after incrementally training the model from task $1$ to $i$, the average accuracy on task $i$ can be calculated by $\frac{1}{i}\sum_{j=1}^{i} a_{i,j}$. We are interested in the final average accuracy after all $n$ tasks have been trained.
	\begin{equation}
	ACC = \frac{1}{n}\sum_{j=1}^{n} a_{n,j}
	\end{equation}
	
	\textbf{Backward Transfer (BT)}: 
	We adopt the forgetting measure in \cite{chaudhry2018riemannian} to calculate the backward transfer.
	\begin{equation}
	\label{Eqn:BT}
	BT = \frac{1}{n-1} \sum_{j=1}^{n-1} [\min_{i\in \{1, 2, ..., n-1\}} a_{n,j} - a_{i,j} ]
	\end{equation}
	If $BT > 0$, positive backward transfer occurs, which means that the following tasks learning helps improve the performance on prior tasks.
	If $BT < 0$, on the other hand, the negative backward transfer causes the performance deterioration on previous tasks. 
	
	\textbf{Forward Adaptation (FA)}:
	The forward adaptation we calculate here is similar with the intransigence measure \cite{chaudhry2018riemannian} and the forward transfer \cite{lopez2017gradient}.
	But we train a randomly initialized model over one task data as the reference model.
	The forward adaptation can be formulated as below:
	\begin{equation}
	FA = \frac{1}{n} \sum_{i=1}^{n} a_{i,i} - a_{i}^{*}
	\end{equation}
	where $a_{i}^{*}$ is the reference model trained from the task $i$ only. 
	We use $a_{i}^{*}$ instead of the joint training accuracy in \cite{chaudhry2018riemannian}. 
	Because $a_{i}^{*}$ is only related with task $i$, we can better understand how the previous tasks learning affects on current task learning.
	For example, if $a_{i,i} - a_{i}^{*} > 0$, it means that the previous tasks knowledge facilitates current task learning (i.e. positive forward adaptation). 

	\subsection{Baselines}
	We apply three state-of-the-art parameter regularization based methods, EWC \cite{kirkpatrick2017overcoming}, PI \cite{zenke2017continual}, and MAS \cite{aljundi2018memory}, to estimate the parameter importance.
	We implement our algorithms MetaCL-$\beta$, MetaCL-$\lambda$ based on their importance estimations, called \{EWC, PI, MAS\}-MetaCL-\{$\beta$, $\lambda$\} (please refer to Appendix B for implementation details).
	We compare them against their original methods (i.e. EWC, PI, MAS) and straightforward fine tune.
	
	\subsection{Results}
	The experiments are conducted on Permuted MNIST, CIFAR-100 and CUB datasets.
	We follow single-head protocol on Permuted MNIST and multi-head protocol on CIFAR-100 and CUB datasets.
	The difference between single-head and multi-head protocol is whether task descriptor is available \cite{chaudhry2018riemannian}.
	For these datasets statistics, please refer to Appendix C.
	We run all methods 3 times and compute the 95\% confidence intervals using the standard deviation across the runs.
	
	
	\begin{table}[h]
		\centering
		\small
		\caption{Experiment results on Permuted MNIST dataset \label{Tab:MNIST}}
		\begin{tabular}{|c|ccc|}
			\hline
			& \multicolumn{3}{|c|}{ Permuted MNIST ($K=20$) } \\
			Method  & ACC (\%) & BT (\%) & FA (\%) \\
			\hline 
			Fine tune	& 46.8 $\pm$ 0.6 & \textbf{-14.8 $\pm$ 1.1} & -0.3 $\pm$ 0.8 \\
			MetaCL, w/o \emph{reg}	& \textbf{48.0 $\pm$ 0.8} & -16.8 $\pm$ 0.8 & \textbf{3.3 $\pm$ 0.8}	\\
			\hline
			MAS	& 55.7 $\pm$ 1.1 & -6.2 $\pm$ 0.9 & 0.2 $\pm$ 0.5 	\\
			MAS-MetaCL-$\beta$	& 56.9 $\pm$ 0.7 & -6.0 $\pm$ 0.7 & 1.3 $\pm$ 0.6 	\\
			MAS-MetaCL-$\lambda$	& \textbf{57.7 $\pm$ 0.6} & \textbf{-5.4 $\pm$ 0.2} & \textbf{1.4 $\pm$ 0.3} \\
			\hline
			PI	& 50.6 $\pm$ 0.9 & -7.2 $\pm$ 1.0 & -4.6 $\pm$ 0.8 	\\
			PI-MetaCL-$\beta$	& 54.1 $\pm$ 0.7 & \textbf{-6.3 $\pm$ 0.4} & -1.9 $\pm$ 0.5 	\\
			PI-MetaCL-$\lambda$	& \textbf{56.0 $\pm$ 0.4} & -7.8 $\pm$ 0.5 & \textbf{2.1 $\pm$ 0.9} 	\\
			\hline
			EWC & 50.4 $\pm$ 0.6 & -12.7 $\pm$ 0.5 & 1.5 $\pm$ 0.4 \\
			EWC-MetaCL-$\beta$ & 53.3 $\pm$ 1.2 & -10.0 $\pm$ 0.5 & 1.8 $\pm$ 0.9 \\
			EWC-MetaCL-$\lambda$ & \textbf{53.8 $\pm$ 0.8} & \textbf{-9.8 $\pm$ 1.0} & \textbf{2.5 $\pm$ 0.1} \\
			\hline
		\end{tabular}
	\end{table}
	
	\begin{table}[h]
		\centering
		\small
		\caption{Experiment results on CIFAR-100 dataset \label{Tab:CIFAR}}
		\begin{tabular}{|c|ccc|}
			\hline
			& \multicolumn{3}{|c|}{ CIFAR-100 ($K=20$) }  \\
			Method  & ACC (\%) & BT (\%) & FA (\%) \\
			\hline 
			Fine tune	& 22.9 $\pm$ 0.8 & -14.0 $\pm$ 0.7 & 4.1 $\pm$ 0.8 	\\
			MetaCL, w/o \emph{reg}	& \textbf{27.5 $\pm$ 1.4} & \textbf{-11.7 $\pm$ 1.3} & \textbf{7.0 $\pm$ 0.6} 	\\
			\hline
			MAS	& 34.2 $\pm$ 0.4 & 1.4 $\pm$ 0.5 & -0.7 $\pm$ 1.0 	\\
			MAS-MetaCL-$\beta$	& 37.0 $\pm$ 0.4 & \textbf{1.9 $\pm$ 0.2} & 1.4 $\pm$ 0.4 	\\
			MAS-MetaCL-$\lambda$	& \textbf{37.5 $\pm$ 0.9} & 1.4 $\pm$ 0.4 & \textbf{2.9 $\pm$ 1.1} 	\\
			\hline
			PI	& 34.8 $\pm$ 0.7 & 1.3 $\pm$ 0.5 & -0.5 $\pm$ 0.9 	\\
			PI-MetaCL-$\beta$	& 39.0 $\pm$ 1.1 & 2.3 $\pm$ 0.8 & 2.8 $\pm$ 0.3 \\
			PI-MetaCL-$\lambda$	& \textbf{39.8 $\pm$ 0.4} & \textbf{2.4 $\pm$ 0.3} & \textbf{4.5 $\pm$ 1.3} 	\\
			\hline
			EWC & 33.5 $\pm$ 0.6 & 1.8 $\pm$ 0.5 & -2.4 $\pm$ 0.9 \\
			EWC-MetaCL-$\beta$ & 37.5 $\pm$ 0.3 & 1.9 $\pm$ 0.2 & 1.2 $\pm$ 0.7  \\
			EWC-MetaCL-$\lambda$ & \textbf{38.3 $\pm$ 0.5} & \textbf{2.1 $\pm$ 0.2} & \textbf{2.0 $\pm$ 0.4}  \\
			\hline
		\end{tabular}
	\end{table}
	
	\begin{table}[h]
		\centering
		\small
		\caption{Experiment results on CUB dataset \label{Tab:CUB}}
		\begin{tabular}{|c|ccc|}
			\hline
			& \multicolumn{3}{|c|}{ CUB ($K=10$) }   \\
			Method  & ACC (\%) & BT (\%) & FA (\%)    \\
			\hline 
			Fune tine & 9.8 $\pm$ 0.8 & \textbf{-32.1 $\pm$ 0.9} & -15.3 $\pm$ 0.7 \\
			MetaCL, w/o \emph{reg} & \textbf{11.4 $\pm$ 0.4} & -36.2 $\pm$ 0.7 & \textbf{-9.1 $\pm$ 0.6} \\
			\hline
			MAS & 26.4 $\pm$ 1.0 & \textbf{-21.7 $\pm$ 1.3} & -7.9 $\pm$ 1.1 \\
			MAS-MetaCL-$\beta$ & 30.4 $\pm$ 1.2 & -22.5 $\pm$ 2.0 & -2.0 $\pm$ 1.3 \\
			MAS-MetaCL-$\lambda$ & \textbf{30.7 $\pm$ 1.2} & -23.6 $\pm$ 1.4 & \textbf{-0.3 $\pm$ 1.2} \\
			\hline
			PI & 38.1 $\pm$ 1.0 & -8.3 $\pm$ 1.3 & -9.6 $\pm$ 1.2 \\
			PI-MetaCL-$\beta$ & 46.1 $\pm$ 1.4 & -3.8 $\pm$ 0.9 & -3.3 $\pm$ 0.6 \\
			PI-MetaCL-$\lambda$ & \textbf{48.7 $\pm$ 1.3} & \textbf{-3.0 $\pm$ 0.7} & \textbf{-2.9 $\pm$ 1.2}  \\
			\hline
			EWC & 32.7 $\pm$ 1.1 & -7.6 $\pm$ 2.6 & -17.4 $\pm$ 2.2  \\
			EWC-MetaCL-$\beta$ & 44.9 $\pm$ 0.2 & -3.1 $\pm$ 0.2 & -8.6 $\pm$ 1.1 \\ 
			EWC-MetaCL-$\lambda$ & \textbf{45.7 $\pm$ 0.7} & \textbf{-2.3 $\pm$ 1.1} & \textbf{-8.5 $\pm$ 1.1} \\
			\hline
		\end{tabular}
	\end{table}
	
	The experiment results on these three datasets are outlined in Tab. \ref{Tab:MNIST}, \ref{Tab:CIFAR}, \ref{Tab:CUB}. 
	Since our algorithms are integrated with various parameter regularization methods, the comparison should be checked within the same regularization method to fairly verify the effectiveness of our methods.
	
	When there is no regularization for knowledge preservation, MetaCL w/o \emph{reg} demonstrates better ACC and stronger forward adaption than straightforward fine tune, with a little cost of BT.
	This demonstrates that MetaCL can exploit the intrinsic features and further strengthen adaptation.
	When we consider parameter regularization, the BT significantly improved.
	For example, in CIFAR-100 dataset (Tab. \ref{Tab:CIFAR}), all MAS, PI and EWC achieve better BT than fine tune (from -14.0\% to 1.5\%). 
	In addition, after applying MetaCL algorithms on these regularization methods, all three metrics ACC, BT and FA are improved.  
	In CUB dataset (Tab. \ref{Tab:CUB}), EWC-MetaCL-{$\beta, \lambda$} outperform original EWC with more than 10\% ACC improvement.
	Finally, compared with the fixed balance strategy MetaCL-$\beta$, the dynamic balance MetaCL-$\lambda$ achieves comparable or better trade-off between BT and FA, and thus further improves ACC.
	
	\textbf{Performance with Different $K$}.
	We evaluate our algorithms on different sizes of training data to comprehensively check the performance.
	The evaluations are conducted on Permutated MNIST and CIFAR-100 with $K=20, 50, 200, 500$, in which $K=20, 50$ can be seen as low-shot scenarios and $K=200, 500$ are standard training. 
	Fig. \ref{Fig:average_acc} shows that the average accuracy changes when more tasks are learned. 
	Tables in Appendix D document all evaluation results. 
	Compared with large size of training data, our algorithms can provide more improvement on low-shot scenarios.
	For example, in CIFAR-100 dataset, PI-MetaCL-{$\beta, \lambda$} outperform original PI with 5\% ACC margin in $K=20, 50$, 3\% in $K=200$ and 1\% in $K=500$.
	This is because standard training procedure could achieve good generalization on large datasets, but it lacks ability to obtain enough intrinsic knowledge from low-shot data.

\begin{figure}[h]
		\centering
		\setlength\tabcolsep{0.5pt}
		\begin{tabular}{cc}
			\includegraphics[width=0.18\textheight]{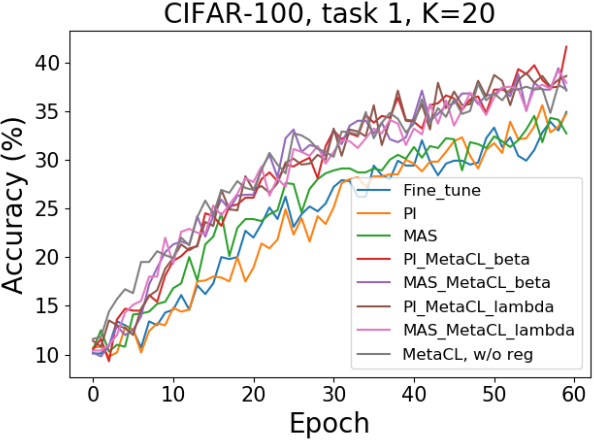} &
			\includegraphics[width=0.18\textheight]{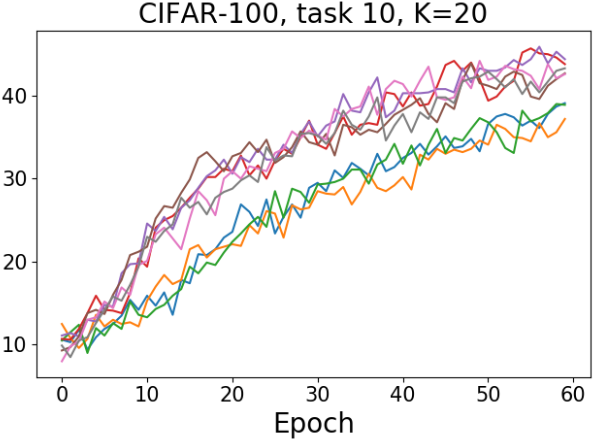}
		\end{tabular}
		\caption{The learning speed and average accuracy comparison among different methods. The MetaCL methods can exploit the intrinsic feature within the limited data and achieve faster learning speed and better model performance. \label{Fig:learning_speed}}
	\end{figure}

	\textbf{Learning Speed Comparison}.
	The MetaCL algorithms not only enhance the forward adaptation but also speed up the learning procedure.
	We run validation on CIFAR-100 testing data every epoch and record the validation accuracy to indicate the learning speed and model performance. 
	Fig. \ref{Fig:learning_speed} illustrates the learning curves when MetaCL algorithm is adopted versus not adopted. 
	The curves of MetaCL methods are always above the original approaches (i.e. orange, blue and green curves), which indicates faster learning speed and higher accuracy. 
	
\begin{figure}[h]
		\centering
		\includegraphics[width=0.37\textheight]{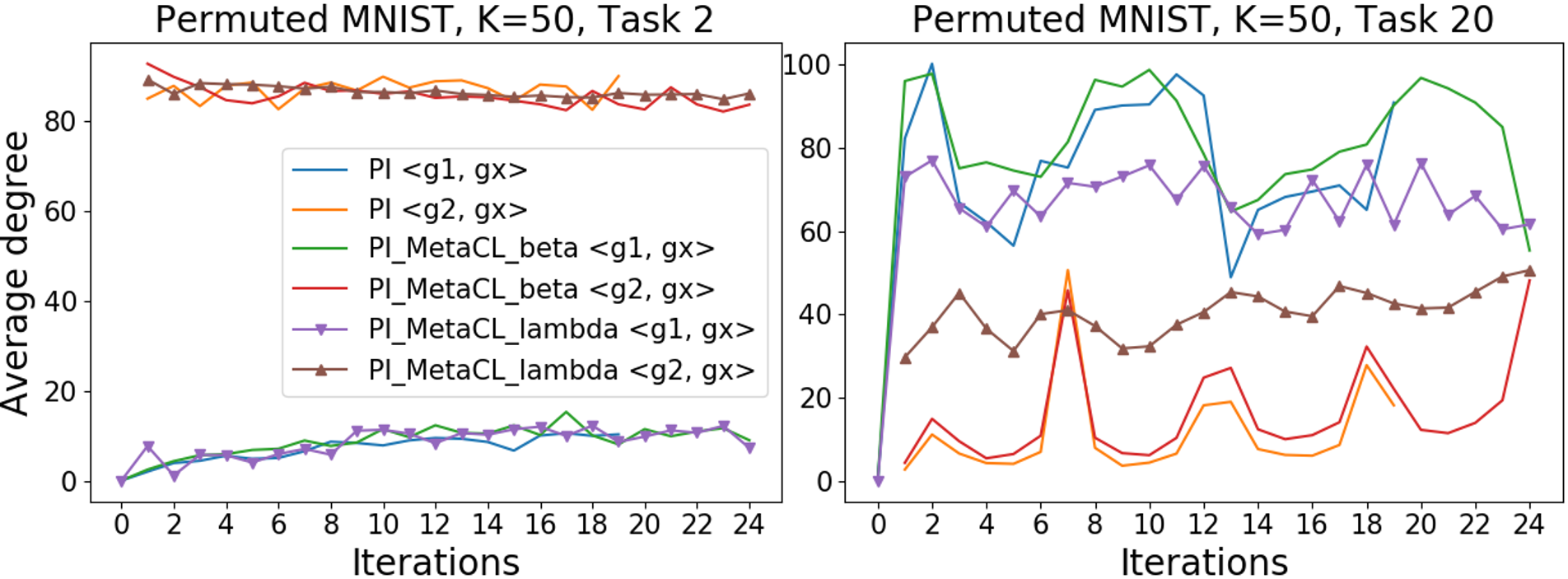} 
		\caption{The gradient angles ($\langle g_1, g_x\rangle$, $\langle g_2, g_x \rangle$) change in optimization procedure.
			The dynamic balance strategy can provide a more stable and better compromise.  \label{Fig:opt_degree}}
	\end{figure}

	\textbf{Regularization Strategy Analysis}.
	With the dynamic balance strategy, the MetaCL-$\lambda$ generally outperforms the fixed balance method.
	On Permuted MNIST with $K=50$, the PI-MetaCL-$\lambda$ surpasses PI-MetaCL-$\beta$ over 8\% in terms of ACC (please check the Table 3 in Appendix D). 
	We take this experiment as an example to analyze the optimization gradients and demonstrate the effectiveness of our new balance strategy.
	As illustrated in Fig. \ref{Fig:opt_degree}, all methods have similar compromise at the beginning (i.e. left figure, learning task 2). 
	But with more tasks learned (right figure), the fixed balance strategy struggles to learn current task 20 (i.e. the angle of $\langle g_1, g_x\rangle$ is big) and cannot provide a stable compromise between current learning object $\ell(\theta)$ and regularization term $\ell^{reg}(\theta)$ (i.e. the angles $\langle g_1, g_x\rangle$, $\langle g_2, g_x \rangle$ are perturbed dramatically).
	As a comparison, the dynamic balance method (purple and grey curves) can give a more stable and better trade-off.
	

	\section{Conclusion}
	In this paper, we generalize the standard continual learning to low-shot scenario. 
	The low-shot setting is more practical and human-like.
	To address the challenges it brings, we develop a new algorithm that can exploit intrinsic features within limited training data and strengthen adaptation ability. 
	To provide a better compromise between learning and forgetting, a new dynamic balance strategy has been proposed. 
	With these two technical components, our algorithm further improve the existing state-of-the-art methods.
	
	In future study, an interesting and more challenging direction is to further decrease the training data (e.g. 1-shot and 5-shot) in continual learning.
	A possible solution for such extreme case is to design new models to further exploit intrinsic information like feature spatial relationship in capsule network \cite{sabour2017dynamic}.


	\bibliographystyle{aaai}
	\bibliography{ref}
	
	\section{Appendix A: The Derivation of Dynamic Balancing Strategy}
	We give the derivation about solving the constrained optimization Eq. \ref{Eqn:Dynamic_balancing_obj_app} in dynamic balance strategy.
	\begin{align}
	\label{Eqn:Dynamic_balancing_obj_app}
	\min_{g_x} \frac{1}{2} \|g_x - g_1 \|^2 \notag \\
	s.t. \quad \langle g_x, g_2 \rangle \ge 0
	\end{align}
	where $g_1, g_2, g_x$ are $p$-dim vectors.
	The Lagrangian dual problem \cite{dorn1960duality} of Eq. \ref{Eqn:Dynamic_balancing_obj_app} is 
	\begin{align}
	\label{Eqn:Lagrangian_dual}
	\max_{\lambda} \min_{g_x} L(\lambda, g_x) = & \frac{1}{2} \|g_x - g_1 \|^2 - \lambda g_2^T g_x \notag \\
	= & \frac{1}{2} g_x^T g_x - g_1^T g_x + \frac{1}{2} g_1^T g_1 - \lambda g_2^T g_x \notag \\
	& s.t. \quad \lambda \ge 0
	\end{align}
	where $\lambda$ is the Lagrange multiplier which is a scalar variable. 
	$\frac{1}{2} g_1^T g_1$ is a constant in this optimization. 
	So we can solve the optimal $g_x$ by calculating $\frac{\partial L(\lambda, g_x)}{\partial g_x}=0$, and get below result:
	\begin{equation}
	\label{Eqn:best_gx_app}
	g_x = g_1 + \lambda g_2
	\end{equation}
	Substitute Eq. \ref{Eqn:best_gx_app} into Eq. \ref{Eqn:Lagrangian_dual}. 
	We have
	\begin{align}
	\label{Eqn:Lagrangian_dual_solved}
	\max_{\lambda} L(\lambda, g_x  = &g_1 + \lambda g_2) = -\frac{1}{2} g_2^Tg_2 \lambda^2 - g_1^Tg_2 \lambda \notag \\
	s.t. \quad &\lambda \ge 0 \notag \\
	& g_x = \lambda g_2 + g_1 
	\end{align}
	Eq. \ref{Eqn:Lagrangian_dual_solved} is a simple one-variable quadratic optimization which is equivalent to Eq. \ref{Eqn:Dynamic_balancing_dual_obj2} in main body.
	\begin{align}
	\label{Eqn:Dynamic_balancing_dual_obj2}
	\min_{\lambda} \frac{1}{2} & g_2^T g_2 \lambda^2 + g_1^T g_2 \lambda  \notag \\
	s.t. \quad & \lambda \ge 0,  \notag \\
	& g_x = \lambda g_2 + g_1 
	\end{align}
	
	\section{Appendix B: Implementation Details}
	Our algorithm can be easily integrated with various network architectures.
	Following the existing literature \cite{zenke2017continual,chaudhry2018riemannian}, the architecture we use in Permuted MNIST dataset \cite{kirkpatrick2017overcoming} is a multi-layer perceptron (MLP) with two hidden layers consisting of 256 units each with ReLU activations.  
	The architecture for CIFAR-100 dataset \cite{krizhevsky2009learning} is illustrated in Tab. \ref{Tab:arch_cifar_100}.
	For dataset CUB \cite{WahCUB_200_2011}, we use the standard ResNet18 \cite{he2016deep} which is pretrained on ImageNet \cite{deng2009imagenet}.
	\begin{table}[h]
		\centering
		\scriptsize
		\caption{The network architecture for CIFAR-100 dataset. $n$ is the number of classes in each task. \label{Tab:arch_cifar_100}}
		\begin{tabular}{cccccc}
			\hline
			Operation & Kernel & Stride & Filters & Dropout & Nonlin. \\
			\hline
			3x32x32 input & & & & & \\
			Conv & 3 $\times$ 3 & $1\times 1$ & 32 & & ReLU \\
			Conv & 3 $\times$ 3 & $1\times 1$ & 32 & & ReLU \\
			MaxPool & & $2\times 2$ & & 0.5 & \\
			Conv & 3 $\times$ 3 & $1\times 1$ & 64 & & ReLU \\
			Conv & 3 $\times$ 3 & $1\times 1$ & 64 & & ReLU \\
			MaxPool & & $2\times 2$ & & 0.5 & \\
			\cline{2-6}
			Task 1: FC & \multicolumn{5}{c}{ $n$ } \\
			\cline{2-6}
			...: FC & \multicolumn{5}{c}{ $n$ } \\
			\cline{2-6}
			Task k: FC & \multicolumn{5}{c}{ $n$ } \\
			\hline
		\end{tabular}
	\end{table}
	
	In all experiments, the sizes of mini-batch and mini-bundle are $100, 10$ respectively. And thus, there are $m=10$ iterations in the inner loop in our MetaCL algorithm. 
	We use SGD with $lr=0.01$ as the inner optimizer and Adam \cite{kingma2014adam} with $lr=0.001$ as the outer optimizer.
	We optimize 5, 20 and 60 epochs for each task on Permutated MNIST, CUB, and CIFAR-100 datasets respectively.  
	
	\section{Appendix C: Dataset Statistics}
	We split original Permuted MNIST, CIFAR and CUB datasets to multiple tasks, and sample first $K$ training images to create low-shot training data.
	The statistics of those datasets are outlined in Tab. \ref{Tab:dataset_statistics}.
	\begin{table*}[!htb]
		\centering
		\caption{Low-shot dataset statistics \label{Tab:dataset_statistics}}
		\begin{tabular}{c|ccc}
			\hline
			Overview & Perm. MNIST & CIFAR & CUB  \\
			\hline
			Num. of tasks & 20 & 10 & 10 \\
			Input size & $1\times 28 \times 28$ & $3 \times 32 \times 32$ & $ 3\times 224 \times 224$ \\
			Evaluation protocol & single-head & multi-head & multi-head \\
			Num. of classes per tasks & 10 & 10 & 20 \\
			Num. of original training images per task & 60000 & 5000 & 600 \\
			Num. of low-shot training images per task & 200 $(K=20)$ & 200 $(K=20)$ & 200 $(K=10)$ \\
			Num. of testing images per task & 10000 & 1000 & 580 \\
			\hline
		\end{tabular}
	\end{table*}
	
	\begin{table*}[h]
		\centering
		\scriptsize
		\setlength\tabcolsep{3.0pt}
		\caption{Experiment results on Permuted MNIST dataset \label{Tab:MNIST_Permutation}}
		\begin{tabular}{|c|ccc|ccc|ccc|ccc|}
			\hline
			& \multicolumn{3}{|c|}{ $K=20$ } & \multicolumn{3}{|c|}{ $K=50$ }  & \multicolumn{3}{|c|}{ $K=200$ } & \multicolumn{3}{|c|}{ $K=500$ }  \\
			Method  & ACC & BT & FA & ACC & BT & FA & ACC & BT & FA & ACC & BT & FA   \\
			\hline 
			Fine tune	& 46.8 $\pm$ 0.6 & \textbf{-14.8 $\pm$ 1.1} & -0.3 $\pm$ 0.8 
			& \textbf{48.7 $\pm$ 1.3} & \textbf{-24.0 $\pm$ 1.5} & -2.3 $\pm$ 0.1 	
			& 47.5 $\pm$ 1.7 & \textbf{-37.4 $\pm$ 1.7} & 0.1 $\pm$ 0.2 	
			& \textbf{49.1 $\pm$ 1.8} & \textbf{-39.7 $\pm$ 1.8} & 1.0 $\pm$ 0.1    \\
			MetaCL, w/o \emph{reg}	& \textbf{48.0 $\pm$ 0.8} & -16.8 $\pm$ 0.8 & \textbf{3.3 $\pm$ 0.8}	
			& 44.3 $\pm$ 0.6 & -32.2 $\pm$ 0.7 & \textbf{1.9 $\pm$ 0.2} 	
			& \textbf{47.8 $\pm$ 0.6} & -38.8 $\pm$ 0.6 & \textbf{1.3 $\pm$ 0.2}	
			& 46.7 $\pm$ 2.2 & -42.5 $\pm$ 2.3 & \textbf{1.6 $\pm$ 0.2}    \\
			\hline
			MAS	& 55.7 $\pm$ 1.1 & -6.2 $\pm$ 0.9 & 0.2 $\pm$ 0.5 	
			& 59.7 $\pm$ 0.5 & \textbf{-7.7 $\pm$ 0.7} & -8.0 $\pm$ 0.3 	
			& 70.4 $\pm$ 0.6 & -8.8 $\pm$ 0.7 & -5.9 $\pm$ 0.3 	
			& 71.8 $\pm$ 0.8 & -8.9 $\pm$ 0.9 & -7.3 $\pm$ 0.0    \\
			MAS-MetaCL-$\beta$	& 56.9 $\pm$ 0.7 & -6.0 $\pm$ 0.7 & 1.3 $\pm$ 0.6 	
			& 61.1 $\pm$ 0.1 & -8.9 $\pm$ 0.2 & -5.2 $\pm$ 0.2 	
			& 67.9 $\pm$ 0.4 & -8.5 $\pm$ 0.3 & -9.3 $\pm$ 0.2 	
			& \textbf{74.1 $\pm$ 0.9} & \textbf{-7.6 $\pm$ 1.1} & \textbf{-6.1 $\pm$ 0.2}    \\
			MAS-MetaCL-$\lambda$	& \textbf{57.7 $\pm$ 0.6} & \textbf{-5.4 $\pm$ 0.2} & \textbf{1.4 $\pm$ 0.3} 	
			& \textbf{62.3 $\pm$ 1.0} & -8.7 $\pm$ 0.8 & \textbf{-4.9 $\pm$ 0.1} 	
			& \textbf{71.2 $\pm$ 1.7} & \textbf{-8.0 $\pm$ 1.7} & \textbf{-4.4 $\pm$ 0.1} 	
			& 72.8 $\pm$ 1.0 & -8.8 $\pm$ 0.8 & -6.4 $\pm$ 0.3    \\
			\hline
			PI	& 50.6 $\pm$ 0.9 & -7.2 $\pm$ 1.0 & -4.6 $\pm$ 0.8 	
			& 51.5 $\pm$ 0.3 & -7.0 $\pm$ 0.6 & -18.3 $\pm$ 0.4 	
			& 66.9 $\pm$ 1.0 & -10.7 $\pm$ 1.0 & -9.3 $\pm$ 0.3 	
			& 66.0 $\pm$ 0.9 & -16.2 $\pm$ 0.7 & -6.1 $\pm$ 0.2    \\
			PI-MetaCL-$\beta$	& 54.1 $\pm$ 0.7 & \textbf{-6.3 $\pm$ 0.4} & -1.9 $\pm$ 0.5 	
			& 55.6 $\pm$ 0.6 & -5.5 $\pm$ 0.1 & -16.0 $\pm$ 0.7 	
			& 65.9 $\pm$ 1.4 & -12.8 $\pm$ 1.0 & -6.9 $\pm$ 0.4 	
			& 60.5 $\pm$ 3.1 & -23.2 $\pm$ 3.4 & -4.6 $\pm$ 0.5    \\
			PI-MetaCL-$\lambda$	& \textbf{56.0 $\pm$ 0.4} & -7.8 $\pm$ 0.5 & \textbf{2.1 $\pm$ 0.9}
			& \textbf{63.8 $\pm$ 0.7} & \textbf{-5.1 $\pm$ 0.3} & \textbf{-5.7 $\pm$ 0.6}	
			& \textbf{73.5 $\pm$ 0.8} & \textbf{-10.1 $\pm$ 0.7} & \textbf{-1.4 $\pm$ 0.2} 
			& \textbf{71.5 $\pm$ 0.1} & \textbf{-14.0 $\pm$ 0.1} & \textbf{-2.6 $\pm$ 0.2}    \\
			\hline
			Joint training (oracle)	& 68.2 $\pm$ 0.2 & - & - 	& 78.2 $\pm$ 0.4 & - & - 	& 88.7 $\pm$ 0.3 & - & - 	& 92.2 $\pm$ 0.2 & - & -    \\
			\hline
		\end{tabular}
	\end{table*}
	
	\begin{table*}[!htb]
		\centering
		\scriptsize
		\setlength\tabcolsep{3.0pt}
		\caption{Experiment results on CIFAR-100 dataset \label{Tab:CIFAR100}}
		\begin{tabular}{|c|ccc|ccc|ccc|ccc|}
			\hline
			& \multicolumn{3}{|c|}{ $K=20$ } & \multicolumn{3}{|c|}{ $K=50$ }  & \multicolumn{3}{|c|}{ $K=200$ } & \multicolumn{3}{|c|}{ $K=500$ }  \\
			Method  & ACC & BT & FA & ACC & BT & FA & ACC & BT & FA & ACC & BT & FA   \\
			\hline 
			Fine tune	& 22.9 $\pm$ 0.8 & -14.0 $\pm$ 0.7 & 4.1 $\pm$ 0.8 	
			& 27.5 $\pm$ 0.9 & -20.0 $\pm$ 1.1 & 4.5 $\pm$ 0.5 	
			& 38.7 $\pm$ 2.6 & -24.0 $\pm$ 2.3 & 5.1 $\pm$ 0.2 	
			& \textbf{40.8 $\pm$ 2.9} & \textbf{-28.7 $\pm$ 2.6} & 3.8 $\pm$ 0.4   \\
			MetaCL, w/o \emph{reg}	& \textbf{27.5 $\pm$ 1.4} & \textbf{-11.7 $\pm$ 1.3} & \textbf{7.0 $\pm$ 0.6} 	
			& \textbf{35.1 $\pm$ 0.6} & \textbf{-15.8 $\pm$ 0.5} & \textbf{8.6 $\pm$ 0.3}
			& \textbf{41.6 $\pm$ 1.4} & \textbf{-23.1 $\pm$ 1.7} & \textbf{7.3 $\pm$ 0.6}	
			& 40.4 $\pm$ 1.9 & -30.2 $\pm$ 2.2 & \textbf{5.0 $\pm$ 0.3}   \\
			\hline
			MAS	& 34.2 $\pm$ 0.4 & 1.4 $\pm$ 0.5 & -0.7 $\pm$ 1.0 	
			& 41.8 $\pm$ 0.5 & 0.9 $\pm$ 0.7 & -3.6 $\pm$ 0.0 	
			& 58.4 $\pm$ 1.2 & 0.9 $\pm$ 0.9 & -1.0 $\pm$ 0.9 	
			& 64.1 $\pm$ 0.7 & 1.8 $\pm$ 0.7 & -4.5 $\pm$ 0.4   \\
			MAS-MetaCL-$\beta$	& 37.0 $\pm$ 0.4 & \textbf{1.9 $\pm$ 0.2} & 1.4 $\pm$ 0.4 	
			& 47.0 $\pm$ 0.7 & \textbf{2.9 $\pm$ 0.6} & 0.2 $\pm$ 0.2 	
			& 61.6 $\pm$ 0.4 & 2.6 $\pm$ 0.5 & 0.6 $\pm$ 0.7 	
			& \textbf{65.3 $\pm$ 0.7} & \textbf{2.1 $\pm$ 0.5} & -3.5 $\pm$ 0.5   \\
			MAS-MetaCL-$\lambda$	& \textbf{37.5 $\pm$ 0.9} & 1.4 $\pm$ 0.4 & \textbf{2.9 $\pm$ 1.1} 	
			& \textbf{48.1 $\pm$ 0.4} & 2.5 $\pm$ 0.4 & \textbf{1.8 $\pm$ 0.8}	
			& \textbf{61.9 $\pm$ 0.8} & \textbf{2.9 $\pm$ 0.6} & \textbf{0.8 $\pm$ 0.6} 	
			& 65.2 $\pm$ 0.9 & 2.0 $\pm$ 0.8 & \textbf{-3.5 $\pm$ 0.2}   \\
			\hline
			PI	& 34.8 $\pm$ 0.7 & 1.3 $\pm$ 0.5 & -0.5 $\pm$ 0.9 	
			& 45.7 $\pm$ 1.3 & 2.9 $\pm$ 0.9 & -1.4 $\pm$ 0.7 	
			& 60.3 $\pm$ 0.5 & 3.7 $\pm$ 0.2 & -3.1 $\pm$ 0.5 	
			& 66.4 $\pm$ 0.3 & 4.6 $\pm$ 0.5 & -5.9 $\pm$ 0.4   \\
			PI-MetaCL-$\beta$	& 39.0 $\pm$ 1.1 & 2.3 $\pm$ 0.8 & 2.8 $\pm$ 0.3 	
			& 50.0 $\pm$ 1.4 & 3.0 $\pm$ 0.5 & 3.1 $\pm$ 1.3 	
			& 62.6 $\pm$ 0.1 & 4.5 $\pm$ 0.3 & -1.4 $\pm$ 0.4 	
			& 66.6 $\pm$ 0.5 & \textbf{4.8 $\pm$ 0.4} & -5.7 $\pm$ 0.5   \\
			PI-MetaCL-$\lambda$	& \textbf{39.8 $\pm$ 0.4} & \textbf{2.4 $\pm$ 0.3} & \textbf{4.5 $\pm$ 1.3} 	
			& \textbf{50.3 $\pm$ 0.3} & \textbf{3.0 $\pm$ 0.3} & \textbf{3.9 $\pm$ 0.3 }
			& \textbf{63.5 $\pm$ 0.2} & \textbf{4.6 $\pm$ 0.2} & \textbf{0.4 $\pm$ 0.5 }
			& \textbf{67.4 $\pm$ 0.5} & 4.5 $\pm$ 0.1 & \textbf{-4.8 $\pm$ 0.2}   \\
			\hline
			Joint training (oracle)	& 40.0 $\pm$ 1.5 & - & -	& 49.8 $\pm$ 1.6 & - & - 	& 60.4 $\pm$ 2.7 & - & - 	& 59.3 $\pm$ 2.4 & - & -   \\
			\hline
		\end{tabular}
	\end{table*}
	
	\section{Appendix D: The Experiment Results with Different Shots}
	We observe the performance changes when we train model on different sizes of training data.
	Tab. \ref{Tab:MNIST_Permutation}, \ref{Tab:CIFAR100} outline all quantitative results when we set $K=20, 50, 200, 500$. 
	The joint training is an oracle strategy in which we feed all tasks data for training in one time.
	
	Compared with Fine tune, MetaCL w/o \emph{reg} always achieves better FA in all circumstances. 
	When we apply parameter regularization, the MetaCL-$\lambda$ generally outperforms other methods in most cases in terms of ACC, BT and FA.
	When compare MetaCL with the original approaches PI and MAS, MetaCL can generally reach higher ACC and FA with comparable BT.
	In some cases like $K=20, 50$ on CIFAR-100, the MetaCL methods achieve better results both on FA and BT. 
	These evidences show that the MetaCL algorithm can effectively strengthen the adaptation ability without sacrificing knowledge preservation.
	In addition, we can also observe that the MetaCL-$\lambda$ can provide a better trade-off between FA and BT compared with MetaCL-$\beta$.
	For example, PI-MetaCL-$\lambda$ significantly surpasses PI-MetaCL-$\beta$ on Permuted MNIST datasets in all three metrics.
	
\end{document}